\documentclass[conference]{IEEEtran}
\IEEEoverridecommandlockouts
\usepackage{cite}
\usepackage{amsmath,amssymb,amsfonts}
\usepackage{algorithmic}
\usepackage{graphicx}
\usepackage{textcomp}
\usepackage{xcolor}
\usepackage[ruled,linesnumbered]{algorithm2e}
\usepackage{multirow}
\usepackage{enumerate}
\usepackage{url}
\usepackage{booktabs}
\usepackage{threeparttable}
\usepackage{endnotes}
\usepackage{bbding}
\usepackage{bm}
\usepackage{CJKutf8}
\usepackage[utf8]{inputenc}
\usepackage{subfigure}
\usepackage{epstopdf}
\usepackage{CJK}

\def\BibTeX{{\rm B\kern-.05em{\sc i\kern-.025em b}\kern-.08em
    T\kern-.1667em\lower.7ex\hbox{E}\kern-.125emX}}

\begin{document}

\begin{CJK*}{UTF8}{gkai}

\title{
Enriching Medcial Terminology Knowledge Bases via Pre-trained Language Model and Graph Convolutional Network
}

\author{\IEEEauthorblockN{Jiaying Zhang$^{1,+}$, Zhixing Zhang$^{1,+}$, Huanhuan Zhang$^1$, Zhiyuan Ma$^{1,*}$, Yangming Zhou$^{1,*}$ and Ping He$^2$}
\IEEEauthorblockA{$^1$School of Information Science and Engineering, East China University of Science and Technology, Shanghai 200237, China \\
$^2$Shanghai Hospital Development Center, Shanghai 200041, China \\
$^*$Corresponding authors\\
$^+$Equal contributor\\
Emails: yuliar3514@gmail.com, ymzhou@ecust.edu.cn\\
}
}

\maketitle

\begin{abstract}
Enriching existing medical terminology knowledge bases (KBs) is an important and never-ending work for clinical research because new terminology alias may be continually added and standard terminologies may be newly renamed. In this paper, we propose a novel automatic terminology enriching approach to supplement a set of terminologies to KBs. Specifically, terminology and entity characters are first fed into pre-trained language model to obtain semantic embedding. The pre-trained model is used again to initialize the terminology and entity representations, then they are further embedded through graph convolutional network to gain structure embedding. Afterwards, both semantic and structure embeddings are combined to measure the relevancy between the terminology and the entity. Finally, the optimal alignment is achieved based on the order of relevancy between the terminology and all the entities in the KB. Experimental results on clinical indicator terminology KB, collected from 38 top-class hospitals of Shanghai Hospital Development Center, show that our proposed approach outperforms baseline methods and can effectively enrich the KB. 
\end{abstract}


\section{Introduction}
\label{Sec:Introduction}
\begin{figure}[!htbp]
\centering
\includegraphics[width=3.5in]{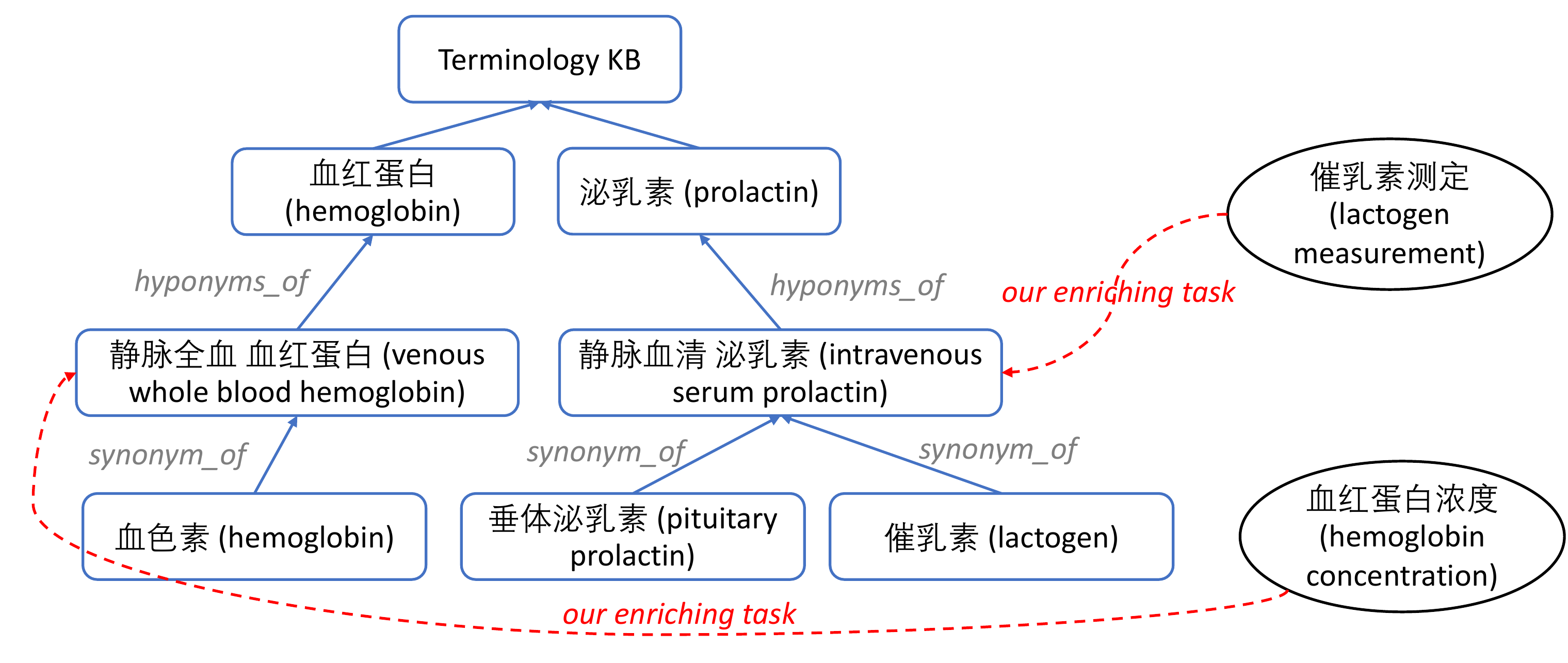}
\caption{An example for our terminology enriching task.}
\label{fig:An example for our terminology enriching task.}
\end{figure}

Recently, terminology knowledge bases (KBs) have attracted increasing attentions and are widely used in clinical domains. However, constructing medical terminology KBs cannot be done once and for all, and terminology enriching (see Fig.~\ref{fig:An example for our terminology enriching task.}) never ends. The enriching is mainly caused by two reasons, namely terminology renaming and synonym adding. The former is common because the standard terminology names are not permanent, they will be replaced by more accurate names over time. For example, in the specimen of venous whole blood, the clinical indicator ``\begin{CJK*}{UTF8}{gkai}血色素\end{CJK*}'' used to be the traditional name of ``\begin{CJK*}{UTF8}{gkai}血红蛋白\end{CJK*}'' (hemoglobin, HGB) in Chinese. The latter owes to the fact that every region, even every hospital, has various names for the same terminology, and it is impossible to incorporate all synonyms into a single KB at once. For instance, collected from different hospitals, in the specimen of venous serum,  the clinical indicator ``\begin{CJK*}{UTF8}{gkai}泌乳素\end{CJK*}'' (prolactin, PRL) may have 7 different synonymous names, namely ``\begin{CJK*}{UTF8}{gkai}催乳素\end{CJK*}''  (lactogen), ``\begin{CJK*}{UTF8}{gkai}垂体泌乳素\end{CJK*}'' (pituitary prolactin), ``\begin{CJK*}{UTF8}{gkai}泌乳素测定\end{CJK*}'' (prolactin measurement), ``\begin{CJK*}{UTF8}{gkai}垂体催乳素\end{CJK*}'' (pituitary lactogen) ``\begin{CJK*}{UTF8}{gkai}催乳素(PRL)\end{CJK*}'' (lactogen PRL), ``\begin{CJK*}{UTF8}{gkai}垂体泌乳素(PRL)\end{CJK*}'' (pituitary prolactin PRL) and ``\begin{CJK*}{UTF8}{gkai}泌乳素(PRL)\end{CJK*}'' (prolactin PRL). In this case, terminology enriching can be considered as a supplementary to existing KBs. 

Existing well-known terminology KBs, such as SNOMED-CT~\cite{donnelly2006snomed}, LONIC~\cite{mcdonald2003loinc} and UMLS~\cite{bodenreider2004unified}, usually enrich terminologies manually to ensure the professional authority, and they have a big team of experts. For example, over 350 individuals are devoted to the original work of SNOMED-CT~\cite{donnelly2006snomed}, and 350 is a large number. Consequently, enriching the existing terminology KBs can not be a timely job, and these well-known KBs are typically enriched and released by few years. To solve the time-consuming and labor-intensive task, designing an automatic terminology enriching method is necessary.

The most relevant work to this task is entity alignment. To achieve the alignment, conventional feature-based methods manually design various features~\cite{wang2013boosting,suchanek2011paris} and embedding-based methods encode KBs into embeddings~\cite{chen2017multilingual,wang2018cross}. However, entity alignment aims to align KBs to other KBs, while terminology enriching need to align a set of terminologies to a terminology KB. It means that the existing entity alignment methods cannot be directly adopted in the enriching task. Moreover, pre-trained language representations, which are popular in many natural language processing (NLP) tasks, have not been employed in existing works. We believe that pre-trained model can further improve our task. Therefore, we try the pre-trained model to hot-start KB embedding and enhance semantic information. 

\begin{figure}[!htbp]
\centering
\includegraphics[width=3.5in]{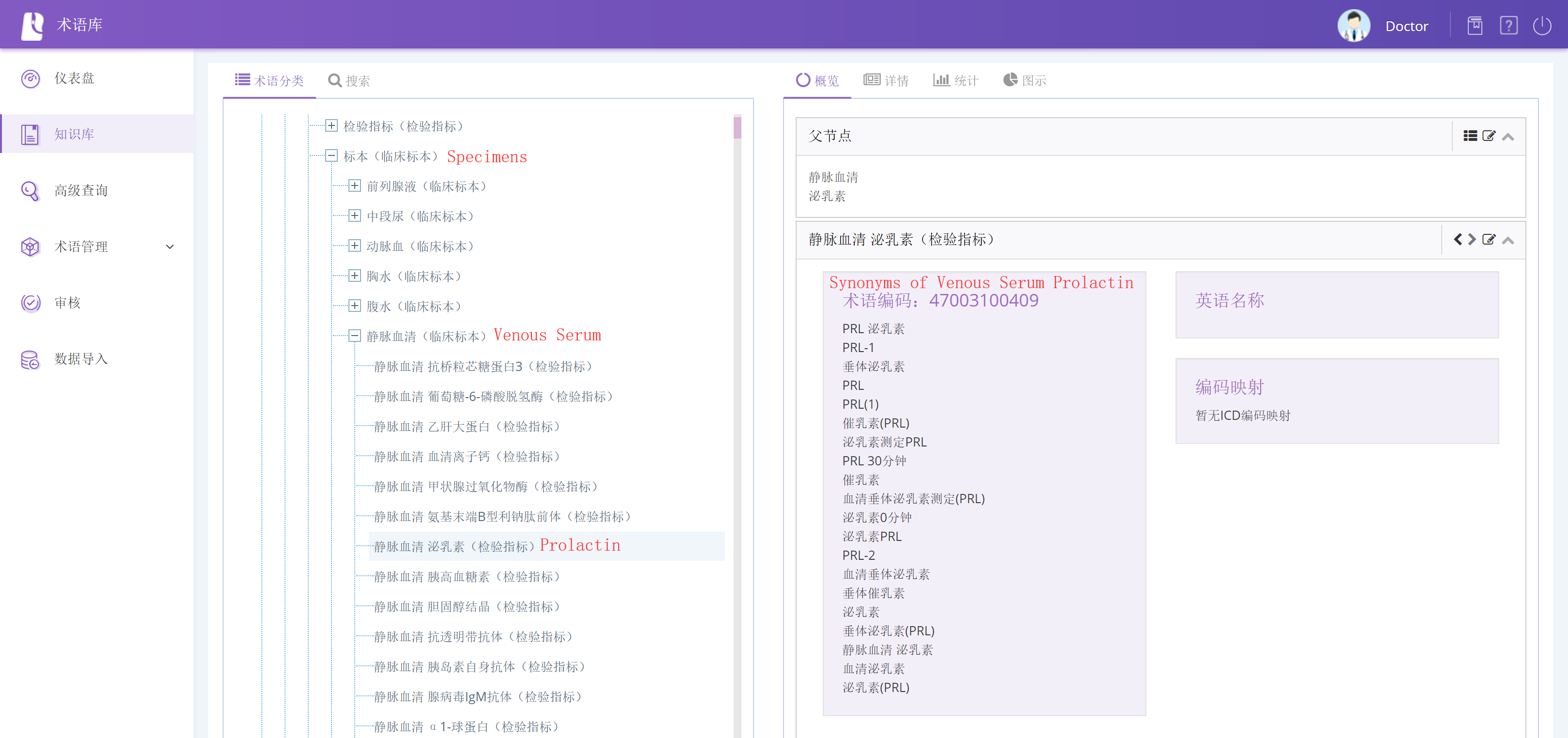}
\caption{The screenshot of the clinical indicator terminology KB.}
\label{fig:The screenshot of the clinical indicator terminology KB.}
\end{figure}

In this paper, we propose a novel terminology enriching model to align a set of terminologies to a terminology KB via a pre-trained language model and graph convolutional network (GCN). Specifically, to predict the relevancy between a candidate terminology and an entity in KB, our model consists of three parts: (1) \textbf{BERT-based Semantic Embedding}. We learn the semantic relevancy between the entity and terminology through the pre-trained language model. (2) \textbf{GCN-based Structure Embedding}. We learn the structure relevancy using GCN. We firstly utilize BERT to initialize the representations of the entity and terminology, respectively, and then further optimize them through GCN. Finally, we calculate the relevancy of the two representations on element-level. (3) \textbf{Embeddings Integration}. We integrate the semantic embedding and the structure embedding for mutual fusion by the multi-Layer perception (MLP) model. The output of the model is the relevancy probability between the entity and terminology. For alignment prediction, we compute the relevancy between the terminology and all the entities in KB, and rank the entities by the relevancy to align. Our experiments are conducted on clinical indicator terminology KB\footnote{We show a demo in http://dcakb.ecustnlplab.com/} (see Fig.~\ref{fig:The screenshot of the clinical indicator terminology KB.}), which contains different clinical indicator names from 38 top-calss hospitals of Shanghai Hospital Development Center. 

The main contributions of this work can be summarized as follows.
\begin{itemize}
    \item We propose a novel terminology enriching method to align a set of terminologies to a terminology KB for the first time. It sequentially integrates BERT-based semantic embedding and GCN-based structure embedding. It is the first time to introduce pre-trained language model to hot-start KB embedding and to enhance semantic information. We also adapt GCN to our enriching task. 
    \item Experimental results show that our proposed model achieves better performance than other baseline methods. In addition, we also show that introducing pre-trained model has a great improvement than pure GCN-based alignment methods.
\end{itemize}

The rest of the paper is organized as follows. Section~\ref{Sec:Related Work} briefly reviews the related work on entity alignment, pre-trained language model and KB embedding. Section~\ref{Sec:Problem Formulation} introduces terminology enriching task. In Section ~\ref{Sec:Proposed Model}, we detail the proposed model. Exerperimental results are described in Section~\ref{Sec:Experimental Analysis}. Finally, the paper is concluded in Section~\ref{Sec:Conclusion and Future Work}.

\section{Related Work}
\label{Sec:Related Work}

\begin{figure*}[!htbp]
\centering
\includegraphics[width=5.0in]{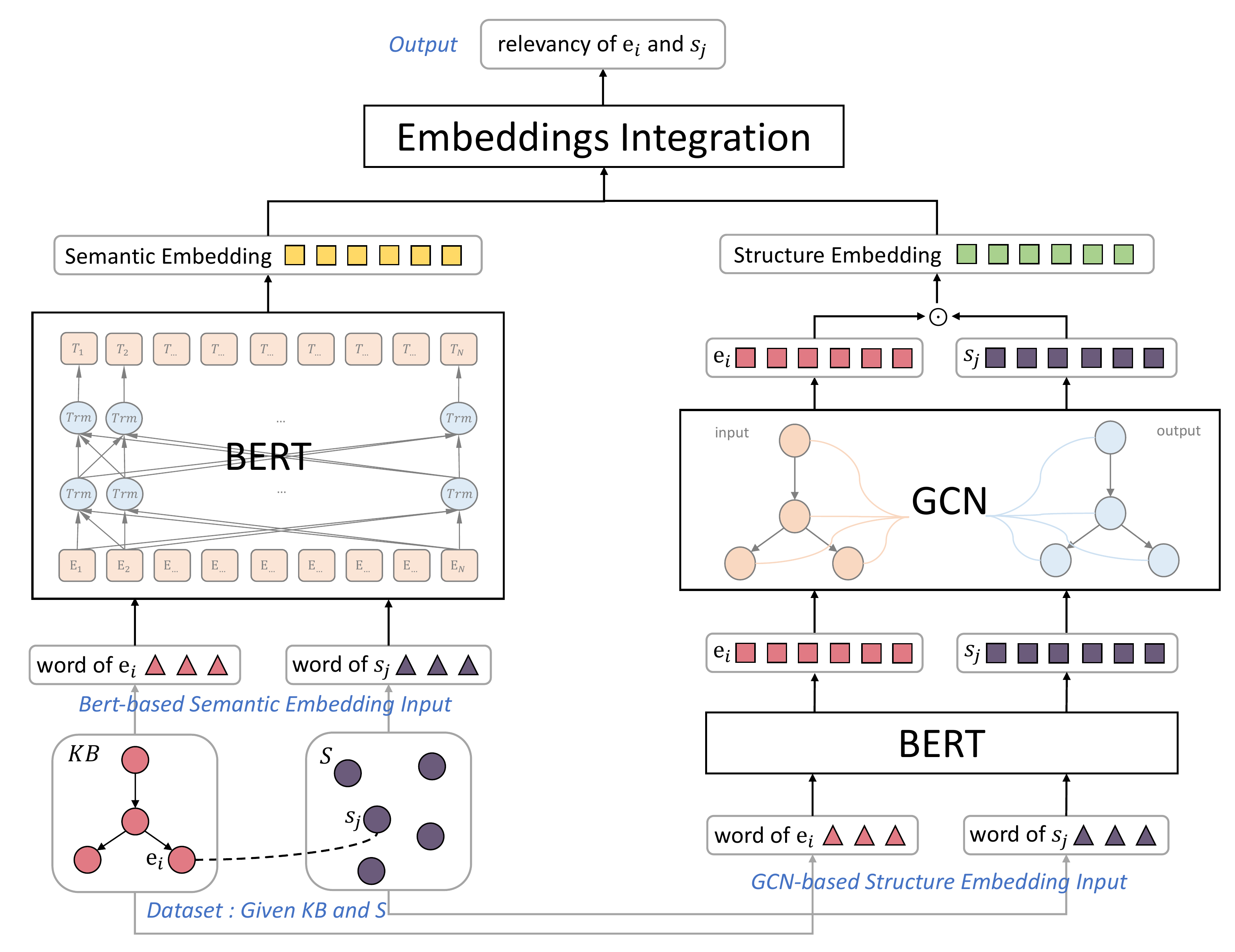}
\caption{Overview of the model for terminology enriching.}
\label{fig:Overview of the model for terminology enriching}
\end{figure*}

\subsection{Entity Alignment}
\label{Sec:Entity Alignment}
The most relevant work to the task of terminology enriching is entity alignment.

Earliest works manually aligned entities. In order to reduce workload, various crowdsourcing algorithms were applied~\cite{wang2015crowd,zheng2017truth,zhuang2017hike} and the alignment quality was promised. Some other conventional feature-based works observe KBs manually and carefully designed various features, such as literal information~\cite{zhang2018effective}, external lexicons~\cite{wang2018using,wang2013boosting} and attribute values~\cite{suchanek2011paris,wang2015effective}. The effectiveness of these methods largely depended on human experience.

Recently, with the emergence of semantic representation learning, many embedding-based methods were proposed, which embedded KBs and achieved the alignment with these embeddings. There existed four basic ideas: MTransE~\cite{chen2017multilingual} encoded KBs in separated embedding spaces and a transformation was learned to align them. JE~\cite{hao2016joint} jointly learned embeddings in a unified space. JAPE~\cite{sun2017cross} introduced attribute embedding in addition to structure embedding. GCN-Align~\cite{wang2018cross} generated entity representation based on neighborhood information and attribute information. Many other works~\cite{sun2018bootstrapping,trsedya2019entity,chen2018co} had been extended according to these four ideas. For example, Zhu et al.~\cite{zhu2017iterative} proposed an iterative and parameter sharing method, which encoded both entities and relations of heterogeneous KBs by TransE and PTransE to obtain knowledge embeddings, and joined these embeddings into a unified semantic space. Zhang et al.~\cite{zhang2019multi} proposed a multi-view embedding method, including name view, relation view and attribute view, and learned their embeddings by Skip-gram, TransE and Convolutional Neural Networks (CNNs) before combining them together for alignment. Pang et al.~\cite{pangiterative} improved GCN-Align by considering both local and global information of attribute representation, incorporated neighbouring attributes as local information, and discarded most frequent attributes as global information.

However, there are three major differences between the above approaches and our task. Firstly, the above methods performed entity alignment between different KBs, but what our enriching task aligns to a terminology KB is a set of terminologies, meaning that we cannot directly adopt the existing methods. Secondly, instead of random initialization, we use a pre-trained language model to hot-start KB embedding. Thirdly, we also employ the pre-trained language model to enhance semantic information.

\subsection{Pre-trained Language Model}
\label{Sec:Pre-trained Language Model}
With the popularity of pre-trained language models in many NLP tasks, the existing methods for applying their pre-trained models to downstream tasks can be divided into two classes: feature-based and fine-tuning. The feature-based methods encoded words into representations and fed these pre-trained representations into tasks-specific architectures as input embeddings, such as ELMo~\cite{peters2018deep}. The fine-tuning methods directly trained their pre-trained models on the downstream tasks and provided task-specific parameters for fine-tuning, such as BERT~\cite{devlin2019bert}, GPT~\cite{radford2018improving}, ERNIE~\cite{zhang2019ernie} and XLNet~\cite{yang2019xlnet}. In this paper, we choose BERT as our pre-trained model for its state-of-art performance.

\subsection{KB Embedding}
\label{Sec:KB Embedding}
KB embedding has been considered as an effective way to encode components of KB including entities and relations into a low-dimensional vector space without losing inherent information~\cite{wang2017knowledge}. TransE~\cite{bordes2013translating} was the most representative method, which interpreted the relation as a translation vector from the head entity to the tail one. TransH~\cite{wang2014knowledge}, TransR~\cite{lin2015learning} and TransD~\cite{ji2015knowledge} were successively proposed to improve TransE in dealing with multi-mapping relations. However, these translation-based embedding models required aligned or shared relations. As another solution, neural-based embedding models were proposed by exploiting deep learning techniques~\cite{zhang2019multi}, such as MLPs~\cite{shi2017proje}, CNNs~\cite{dettmers2018convolutional}, and GCNs~\cite{schlichtkrull2018modeling}. In this paper, we use GCN for KB embedding.

\section{Problem Formulation}
\label{Sec:Problem Formulation}
Formally, the existing terminology knowledge base is defined as $KB = (E, R)$, where $E$ denotes the set of entities (i.e. terminologies) and $R$ denotes the set of relations between entities. Each knowledge can be described by one of the following two triples: $T^{Syn} = \{(h, r, t) | h,t \in E, r = synonymous\_of \in R\}$ and $T^{Hyp} = \{(h, r, t) | h,t \in E, r = hyponyms\_of \in R\}$. For example, in the specimen of venous serum, (\begin{CJK*}{UTF8}{gkai}催乳素\end{CJK*}, $synonym\_of$, \begin{CJK*}{UTF8}{gkai}静脉血清泌乳素\end{CJK*}) means that the clinical indicator ``\begin{CJK*}{UTF8}{gkai}催乳素\end{CJK*}'' (lactogen) is a synonym of ``\begin{CJK*}{UTF8}{gkai}静脉血清泌乳素\end{CJK*}'' (intravenous serum prolactin), and (\begin{CJK*}{UTF8}{gkai}泌乳素\end{CJK*}, $hyponyms\_of$, \begin{CJK*}{UTF8}{gkai}静脉血清 泌乳素\end{CJK*}) means that ``\begin{CJK*}{UTF8}{gkai}静脉血清 泌乳素\end{CJK*}''(intravenous serum prolactin) belongs to ``\begin{CJK*}{UTF8}{gkai}泌乳素\end{CJK*}''.

Given existing terminology knowledge base $KB$ and a set of candidate terminologies $S = \{s_1, s_2, \dots, s_m\}$ to be updated, where $m$ is the number of terminologies, the task is to automatically pick the \textit{synonymous} pair set $P = \{(e_i, s_j) | e_i \in E, s_j \in S\}$ and align $s_j$ to $e_i$ respectively.

\section{Proposed Model}
\label{Sec:Proposed Model}
In this section, we present our proposed model for terminology enriching. Given the terminology KB and a candidate terminology $s_j$, each entity in KB is sequentially extracted to calculate a relevancy with $s_j$, and then the optimal alignment results are ranked by relevancy.

The overview model to obtain the relevancy of an entity $e_i$ in KB and $s_j$ is shown in Fig.~\ref{fig:Overview of the model for terminology enriching}, in which the relevancy is computed by integrating semantic information and structural information into MLP. To get the semantic embedding, pre-trained language model BERT is utilized. To get the structure embedding, GCN is used. The representations of $e_i$ and $s_j$ are initialized by BERT, and further embedded through GCN respectively. The relevancy embedding with structural information is obtained by the dot production of the two representations. Each stage is trained separately and sequentially. 

\subsection{BERT-based Semantic Embedding}
\label{Sec:BERT-based Semantic Embedding}
\begin{figure}[!htbp]
\centering
\includegraphics[width=3.0in]{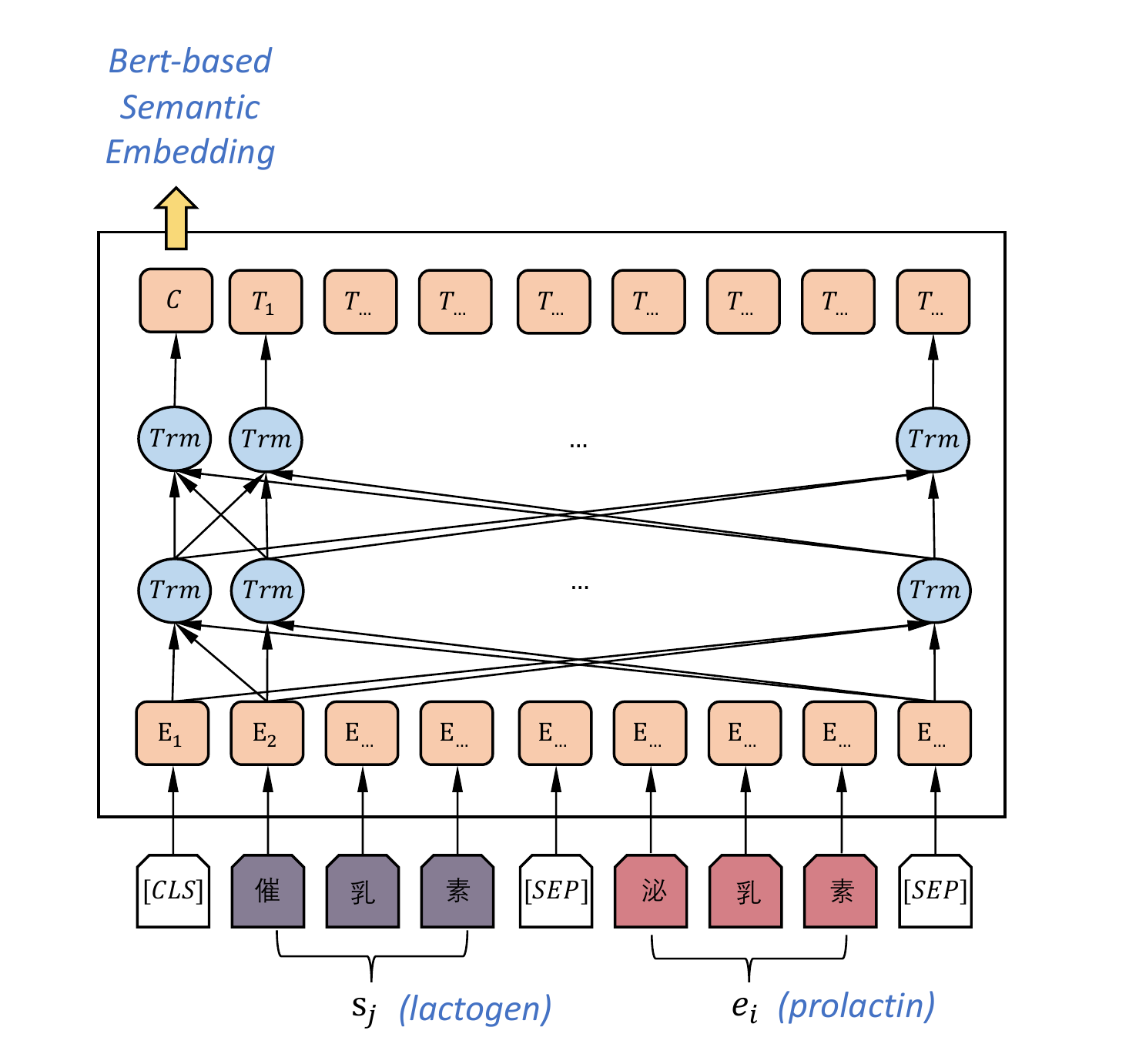}
\caption{BERT model structure for semantic embedding. Note that the figure is used to show the model structure, the entity ``静脉血清泌乳素'' (intravenous serum prolactin) is replaced by the ``泌乳素'' (prolactin) for the drawing reason.}
\label{fig:BERT model structure for semantic embedding.}
\end{figure}

Intuitively, synonymous terminologies are assumed to share similar semantics in context. Based on the assumption, pre-trained language representation models have been proved to be effective in capturing semantic relevancy and work as a routine component in many NLP tasks~\cite{zhang2019ernie}. In this paper, BERT~\cite{devlin2019bert} is chosen to represent semantic embedding, the structure of which are shown in Fig.~\ref{fig:BERT model structure for semantic embedding.}.

\textbf{BERT Embedding Model.}
To be specific, given an entity $e_i$ in the terminology KB and a candidate terminology $s_j$ in S, these two inputs are merged into a sequence \{[CLS] $x$ [SEP] $y$ [SEP]\}, where $x = s_j$, $y = e_i$, and they are word-level tokenized. For each token in the sequence, its corresponding token embedding, segment embedding and positional embedding are summed as the input embedding. 

The BERT model consists of multiple bidirectional Transformer encoder layers. The output of the $l$-th layer is the input of the $l+1$-th layer, and the semantic embedding is computed as follows:
\begin{align}
    \Tilde{\bm{H}}^{(l)} &= LayerNorm(\bm{H}^{(l)} + MhAtt(\bm{H}^{(l)}))\\
    \bm{H}^{(l+1)} &= LayerNorm(\Tilde{\bm{H}}^{(l)} + FFN( [\Tilde{\bm{H}}^{(l)}]_+))
\end{align}
where MhAtt($\cdot$) is the multi-head self-attention~\cite{vaswani2017attention}, LayerNorm($\cdot$) is the layer normalization~\cite{ba2016layer} and $[x]_+=\max \{0, x\}$ represents the maximum between 0 and $x$.

Then, the [CLS] token embedding is taken as BERT-based semantic embedding, which represents the relevancy of $e_i$ and $s_j$. 

\textbf{Loss Function.}
To fine-tune BERT, we use a set of known \textit{synonymous} pair set $P^+$, and $P^-$ stands for the negative sample set of $P^+$. If $(e_i, s_j) \in P^+$ (or $(e_i^{'},s_j^{'}) \in P^-$), we define $y_{se}$ = 1 (or 0). Then we put BERT-based semantic embedding into the sigmoid activation function to get $\hat{y}_{se}$.

When training the model, the optimizing objective is to minimize the binary cross-entropy function:
\begin{align}
    L_{se} &= -y_{se}\log(\hat{y}_{se})-(1-y_{se})\log(1-\hat{y}_{se})
\end{align}

\subsection{GCN-based Structure Embedding}
\label{Sec:GCN-based Structure Embedding}
BERT-based semantic embedding model only utilize entity information without considering the structure of whole graph. However, relations in a KB usually offer additional guidelines for the model and synonyms are non-ignorable in a terminology KB. GCNs are a type of neural network that can produce a node-level embedding using the structure of a graph and embedding of nodes, so that neighborhood information such as synonym embedding and hyponymy relation can be encoded through an end-to-end learning approach. 

\textbf{GCN Embedding Model.}
Note that synonyms describe entities alternatively and hyponymy relation offers categorical information of the entity. GCN is able to combine these information and projects the entities into the same continuous vector space where entities and their synonyms are close to each other. 

Our GCN model consists of $L$ GCN layers, and the $l$-th GCN layer has two inputs. One is an $n\times d^{(l)}$ node embedding matrix $\bm{H}^{(l)}$, where $n$ is the number of nodes and $d^{(l)}$ represents the node embedding dimension of the $l$-th layer. The other is the adjacent matrix $\bm{A}$  which is an $n\times n$ matrix. The $l$-th layer uses the following normalized graph Laplacian transformation\cite{belkin2002laplacian} to obtain the output.
\begin{align}
    \bm{H}^{(l+1)} &= \sigma (\hat{\bm{D}}^{-\frac{1}{2}}\hat{\bm{A}}\hat{\bm{D}}^{-\frac{1}{2}}\bm{H}^{(l)}\bm{W}^{(l)})
\end{align}
where $\sigma$(·) is the non-linear activation function, \bm{$A$} is the $n\times n$ adjacent matrix describing the connectivity of the KG, and $\hat{\bm{A}}=\bm{A}+\bm{I}$  where \bm{$I$} represents the identity matrix. \bm{$\hat{D}$} is a diagonal matrix with entries $\bm{\hat{D}}_{ii}=\Sigma_j\bm{\hat{A}}_{ij}$. $\bm{W}^{(l)}$ is the weight matrix of the $l$-th layer.

\textbf{Node Embedding.}
In the GCN embedding model, $\bm{H}^{(0)}$ is initialized by the node embedding generated by BERT. Specifically, BERT is firstly fine-tuned using the aforementioned method, then we transformed the entities $e_i$ and candidate $s_j$ terminologies into the specific sequence \{[CLS] $x$ [SEP][SEP]\}, i.e. we set $x=e_i$, $x=s_j$ separately. Then we utilize the above-mentioned model to compute the embedding of every word. Note that entities and candidate terminologies are both treated as nodes in GCN model for GCN can only handle node inputs, where candidate terminologies can be seen as isolated nodes in graph. Finally the token embedding of [CLS] is taken out as node embedding, since $y$ in the aforementioned sequence is set empty, it only contains information of $x$.

\textbf{Adjacent Matrix.}
There are two types of adjacent matrices for entities and the terminologies, respectively. 

Unlike Wang et al.~\cite{wang2018cross} designed particular connectivity matrix, for entities in terminology KB, we simply set entries in adjacent matrix $\bm{A}_{ij}$ to 1 when an edge from entity $e_j$ to entity $e_i$ exists. The reason is, there are two relations, namely ‘synonym of’ and ‘hyponyms of’, which we believe deliver important information from entity $e_j$ to entity $e_i$ equally. 

For candidates terminologies, we use an all-zero matrix as adjacent matrix, indicating no edge exists between candidates terminologies. 

\textbf{Relevancy Embedding with Structural Information.}
The output of the $L$-th GCN layer are the node embeddings of entities and candidate terminologies. We then proceed with an element-wise multiply operation on these two node embeddings and obtain GCN-based structure embedding, which is the relevancy embedding with structural information.

\begin{figure}[!htbp]
\centering
\includegraphics[width=3.0in]{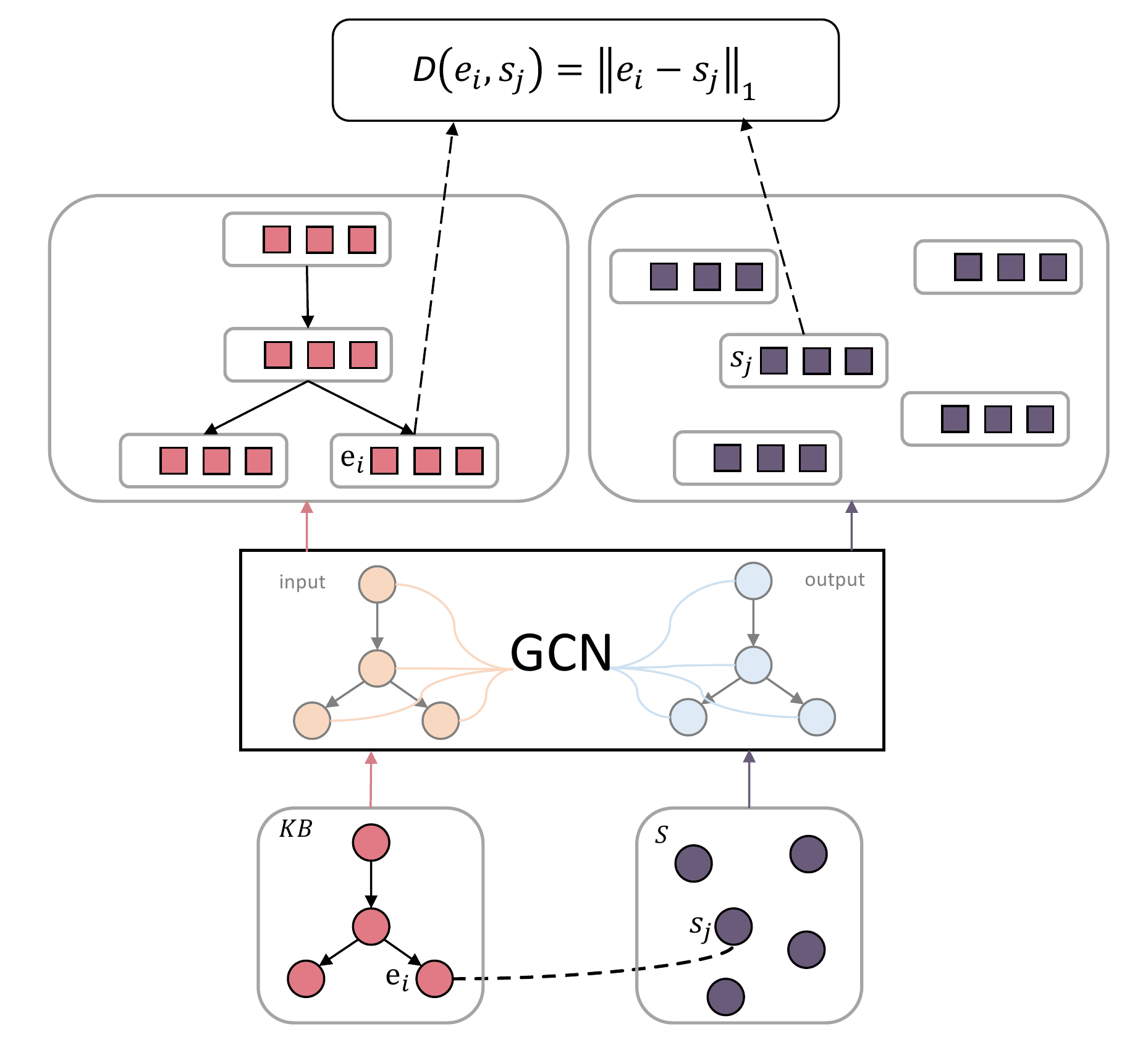}
\caption{GCN model structure for structure embedding.}
\label{fig:GCN model structure for structure embedding.}
\end{figure}

\textbf{Loss Function.}
In order to project the node embedding into the vector space where entities and their corresponding candidate terminologies are close to each other, we utilize margin-based distance loss functions to optimize the problem. The distance definition are shown in Fig.~\ref{fig:GCN model structure for structure embedding.}, pairs of entity node embeddings and candidate terminology node embeddings are taken out to calculate the distance. We define the $n$-th moment distance function as:
\begin{align}
    D(e_i,s_j)=\left\|e_i-s_j\right\|_n
\end{align}
And the loss function is:
\begin{align}
    L=\sum_{(e_i,s_j)\in P^+,(e_i^{'},s_j^{'})\in P^-}{[D(e_i,s_j)+\gamma-D(e_i^{'},s_j^{'})]_+}
\end{align}
where $\gamma>0$ is the hyper-parameter which represents the margin between positive samples and negative samples. We adopt Adam~\cite{kingma2014adam} to minimize the loss function, in order to minimize the distance between positive pair while maximizing the distance between negative pair. 

\subsection{Embeddings Integration}
\label{Sec:Embeddings Integration}
Both the semantic relevancy and the structure relevancy contain important information for alignment, an MLP model is adopted for mutual fusion of the semantic embedding $\bm{X}_{se}$ and the structure embedding $\bm{X}_{st}$. 

\textbf{MLP Model.} 
The output of the hidden layers are represented as follows:
\begin{equation}
\bm{H}^{(0)} = [\bm{X}_{se};\bm{X}_{st}]\\
\end{equation}
\begin{equation}
\hat{\bm{H}}^{(l)} = \sigma(\bm{W}^{(l)}\bm{H}^{(l)}+\bm{b}^{(l)})\\
\end{equation}
\begin{equation}
\bm{H}^{(l+1)} = \sigma(\hat{\bm{W}}^{(l)}\hat{\bm{H}}^{(l)}+\hat{\bm{b}}^{(l)})\\
\end{equation}
\begin{equation}
\bm{H}^{(f)} = sigmoid(\bm{W}^{(f-1)}\bm{H}^{(f-1)}+\bm{b}^{(f-1)})
\end{equation}

where $\bm{H}^{(0)}$ is the input and $\bm{H}^{(f)}$ is the output.

\textbf{Loss Function.}
Similar to BERT, if $(e_i, s_j) \in P^+$ (or $(e_i^{'},s_j^{'}) \in P^-$), $y_{in}$ = 1 (or 0) and $\hat{y}_{in}$ = $\bm{H}^{(f)}$. The loss function is still the binary cross-entropy function:
\begin{align}
    L_{in} &= -y_{in}\log(\hat{y}_{in})-(1-y_{in})\log(1-\hat{y}_{in})
\end{align}

\section{Experimental Analysis}
\label{Sec:Experimental Analysis}
In this section, to evaluate the effectiveness of our model, we compare our method with basic methods, feature-based methods and embedding-based methods. To evaluate the importance of different components in our model, we vary our model in different ways, including ablation analysis, varying data size and model hyper-parameters, measuring the changes in performance of terminology enriching. 

\subsection{Dataset}
\label{Sec:Dataset}
Clinical indicator terminology KB~\cite{zhang2018effective}, which contains different clinical indicator names from 38 top-class hospitals of Shanghai Hospital Development Center, are included in our experiments. The terminology KB contains 15,960 entities, including 3,636 standard terminology names and 12,324 terminology synonyms, and 17,930 relations including 5,606 hyponymy relations and 12,324 description relations. 


Table~\ref{table:Statistics of the KB and the candidate terminologies} outlines the detail statistics of our dataset. We choose 743 standard terminology names which contains at least three synonyms so that we can take full advantage of structure embedding. Subsequently, we split the corresponding synonyms into two parts, namely KB structure set and candidate set in ratio of 2:8. We take KB structure set as entities in graph. Candidate set is removed from KB and treated as the set of candidate terminologies $S$. We further split the candidate set into training set, validation set and test set in ratio of 3:2:5. The KB we used in our experiment has 2,554 nodes, and the number of standard terminology names is 743. In total, there are 1,343 candidate terminologies in the training set, 895 in the validation set and 2,239 in the test set. 

\begin{table}[!ht]
\begin{center}
\caption{Statistics of the KB and the candidate terminologies}
\label{table:Statistics of the KB and the candidate terminologies}
\begin{tabular}{l|lr}
\toprule
Dataset                                                       & Type                 & Total \\ \midrule
\multirow{4}{*}{KB}                                                               & Standard Terminology & 743   \\
                                                                                  & Terminology Synonym  & 1,212  \\
                                                                                  & Hyponyms Relation    & 1,486  \\
                                                                                  & Synonym Relation     & 1,212  \\ \midrule
\multirow{3}{*}{\begin{tabular}[c]{@{}l@{}}Candidate \\ Terminology\end{tabular}} & Training Set         & 1,343  \\
                                                                                  & Test Set             & 2,239  \\
                                                                                  & Validation Set       & 895   \\ \bottomrule
\end{tabular}
\end{center}
\end{table}

\subsection{Experiment Settings}
\label{Sec:Experiment Settings}
The model is implemented using Keras~\cite{chollet2015keras} with TensorFlow~\cite{abadi2016tensorflow} backend run on NVIDIA GeForce GTX 1080Ti GPU. All three parts in the model are optimized by Adam. Due to the high cost of pre-train BERT and lack of large scale corpus, we directly adapt parameters pre-trained in Chinese by Google. During fine-tuning process, we lock first 11 layer and train the last layer only. BERT is optimized in learning rate of $5\times10^{-5}$. 
Learning rate of GCN model is $1\times10^{-5}$, layer output dimension is set to 768. The embedding integration model is optimized in learning rate of $1\times10^{-4}$. 

We use \textit{Hit@k}, which is the percentage of properly aligned entities ranked in the top k candidates, as metric, and we take $k$ in values of 1, 5, 10.

\subsection{Comparison with State-of-the-art Methods}
\label{Sec:Comparison with State-of-the-art Methods}
We firstly compare our proposed model with one basic method and four state-of-the-art methods published in the last year. The basic method firstly preprocess the data, including special character replacement and abbreviation separation before utilizing longest common subsequence threshold to filter synonyms. Besides the basic method, the state-of-the-art methods tried to design feature vectors or obtain KB embeddings. For example, Zhang et al.~\cite{zhang2018effective} combined different character similarity algorithms (e.g. cosine similarity), and trained a binary classifier to find synonyms. Wang et al.~\cite{wang2018using} used a knowledge graph for both hypernymy and synonym detection between Chinese symptoms. Wang et al.~\cite{wang2018cross} generated entity representation and attribute representation based on neighborhood information. Pang et al.~\cite{pangiterative} improved Wang et al.~\cite{wang2018cross} by considering both local and global information of attribute representation. Note that there is neither attribute information nor external lexicons in our data, these existing methods are adjusted and then applied to the experiments.

\begin{table}[!ht]
\begin{center}
\caption{Experimental Results of State-of-the-art Methods and Our Proposed Model}
\label{table:Experimental Results of State-of-the-art Methods and Our Proposed Model}
\setlength{\tabcolsep}{3.5mm}{ 
\begin{threeparttable}
\begin{tabular}{@{}l|ccc@{}}
\hline
\toprule
Methods                                                           & Hits@1 & Hits@5 & Hits@10 \\ \midrule
Basic method                                                      & 20.10  & 50.92  & 63,96   \\ 
Zhang et al.~\cite{zhang2018effective}                            & 45.42  & 74.90  & 80.88   \\ 
Wang et al.~\cite{wang2018using}                                  & 14.60  & 17.95  & 18.22   \\ 
\begin{tabular}[c]{@{}l@{}}Wang et al.~\cite{wang2018cross}$^\star$\\ Pang et al.~\cite{pangiterative}$^\star$\end{tabular} & 31.76  & 53.68 & 61.81  \\ 
\textbf{Our model}                                                         &\textbf{59.58}&\textbf{84.01}&\textbf{87.63}\\ \bottomrule
\end{tabular}
\begin{tablenotes}
 \item[$^\star$] Since Pang et al.~\cite{pangiterative} improved the attribute embedding, which is missing in our data, of Wang et al.~\cite{wang2018cross}, these two methods share the same scores.
\end{tablenotes}
\end{threeparttable}}
\end{center}
\end{table}

The experimental results are shown in Table~\ref{table:Experimental Results of State-of-the-art Methods and Our Proposed Model}. From this table, we clearly observe that our model achieves much higher scores among all these reference algorithms with the Hits@1 of 59.58\%, the Hits@5 of 84.01\% and the Hits@10 of 87.63\%, which means the combination of the pre-trained model and GCN in our model is complementary. The introduction of pre-trained representations can make full use of both the unsupervised pre-training and supervised training data for better enriching. GCN can effectively utilize the structural characteristics of KB and integrate the neighbor information of the entity into its own representation. Meanwhile, among all baselines, Zhang et al.~\cite{zhang2018effective} is the strongest one while Wang et al.~\cite{wang2018using} is the worst. The reason for the phenomenon is that a small amount of training data is enough for literal similarity calculated by Zhang et al.~\cite{zhang2018effective}, and the performance of Wang et al.~\cite{wang2018using} depends largely on their knowledge graph, which does not cover enough required knowledge in the experiments. 

\subsection{Ablation Analysis}
\label{Sec:Ablation Analysis}
To investigate the importance of model components, we explore the effects of the BERT and GCN for our model. In addition, we also study the effects of different initialization of GCN representations in our model.  

\begin{table}[!ht]
\begin{center}
\caption{Comparative Results for Ablation Analysis of Our Proposed Model}
\label{table:Comparative Results for Ablation Analysis of Our Proposed Model}
\setlength{\tabcolsep}{1mm}{ 
\begin{threeparttable}
\begin{tabular}{@{}l|ccc@{}}
\hline
\toprule
Components                         & Hits@1 & Hits@5 & Hits@10 \\ \midrule
\textbf{Our model}                 & \textbf{59.58}  & \textbf{84.01}  & \textbf{87.63}   \\ 
- w/o BERT$^\star$                           & 40.24  & 68.11  & 72.85   \\ 
- w/o GCN$^\star$                            & 52.88  & 81.60  & 86.69   \\ 
- GCN w/o initialized by BERT$^\circ$             & 49.04  & 81.69  & 86.78   \\ 
- GCN initialized by BERT w/o fine-tuning$^\circ$ & 56.41  & 83.11  & 87.32   \\ \bottomrule
\end{tabular}
\begin{tablenotes}
 \item[$^\star$] \textbf{w/o BERT} and \textbf{w/o GCN} refer to our model without BERT as semantic embedding and GCN as structure embedding respectively.
 \item[$^\circ$] \textbf{GCN w/o initialized by BERT} refers to initialization of GCN representations is generated randomly without BERT. \textbf{GCN initialized by BERT w/o fine-tuning} refers to that BERT is used to initialize GCN representation, but fine-tuning process is omitted and original BERT pre-training parameters remain unchanged.
\end{tablenotes}
\end{threeparttable}}
\end{center}
\end{table}

As demonstrated in Table~\ref{table:Comparative Results for Ablation Analysis of Our Proposed Model}, we have the following observations: (1) Both BERT and GCN play important roles in our model. The pre-trained language model BERT show its ability to disambiguate, and can improve Hits@k score by more than 15\%. GCN model can capture structural information contained in the graph, so that it can help to improve Hits@1 score to 59.58\%. (2) Since the most nodes in terminology KB are synonyms and lack of distinctive relations, BERT node embedding without fine-tuning can improve Hits@1 score from random initialized node embedding from 49.04\% to 56.41\%, which is more than 7\%, and specific fine-tuning can further improve Hits@1 score of our proposed model by more than 3\%.

\subsection{Impacts of Different Sizes of Training Data}
\label{Sec:Impacts of Different Sizes of Training Data}
\begin{figure}[!htbp]
\centering
\includegraphics[width=3.6in]{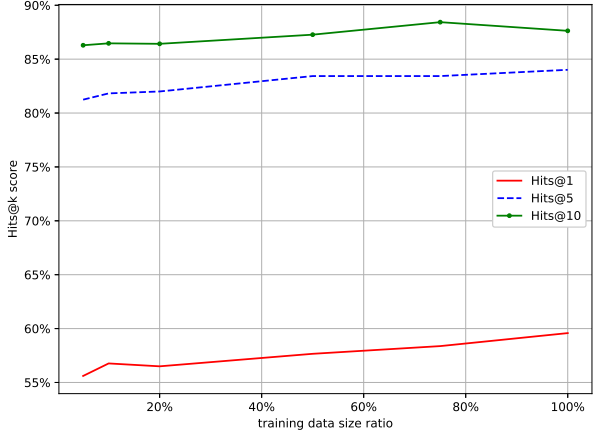}
\caption{Hit@k score in different data size.}
\label{fig:Hit@k score in different data size}
\end{figure}

To study how the size of training set influences the performance and test the scalability of our model, we use different proportions of the training data and calculate the Hits@k scores. Intuitively, the more training data are used, the better results can be obtained. In this paper, considering the total number of training data and the cost of time, we pick 5\%, 10\%, 20\%, 50\%, 75\% of the training data and summarize the comparative results in Fig.~\ref{fig:Hit@k score in different data size}. 

From the results of different data size in Fig.~\ref{fig:Hit@k score in different data size} we can observe that our model performs better as the training data size increases, although the score improves slightly and the increasing rate is low. The results show that the model can perform well in small training data. This is reasonable because our model uses pre-trained representations to enhance semantic embedding and hot-start structure embedding. Additionally, note that our test set is much larger than training set, it also proves that our model can enrich the terminology KB with a small amount of labeled training data. 

\subsection{Impacts of Different Graph Convolutional Networks}
\label{Sec:Impacts of Different Graph Convolutional Networks}
\subsubsection{Impacts of Different GCN Layers}
We investigate the influence of different GCN layer number by running the proposed model on layer numbers $L$ of 1, 2 and 3. Specifically, GCN layers except the last layer use activation function $[\cdot]_+$, and the last layer uses no activation function to output node embedding. Table~\ref{table:Comparative Results for Different GCN layers} presents the comparative results. 

\begin{table}[!ht]
\begin{center}
\caption{Comparative Results for Different GCN layers}
\label{table:Comparative Results for Different GCN layers}
\begin{tabular}{@{}l|ccc@{}}
\toprule
GCN Layers    & Hits@1 & Hits@5 & Hits@10 \\ \midrule
$L=1$  & \textbf{59.58} & \textbf{84.01} & \textbf{87.63}  \\
$L=2$  & 56.19 & 82.85 & 86.91  \\
$L=3$  & 54.35 & 82.80 & 86.55  \\ \bottomrule
\end{tabular}
\end{center}
\end{table}

From Table~\ref{table:Comparative Results for Different GCN layers}, we can observe that our proposed model gets the best performance when GCN layer number $L=1$, which owes to the fact that complex GCN model will overfit the train data while model is hot-started by pre-trained language model. Therefore, we set hyper-parameter $L=1$ in the rest of our experiments.

\subsubsection{Impacts of Different GCN Loss Functions}
To further analyze the impacts of the margin and the distance function used in the GCN loss function, we compare the performances of models with different parameter values. Experimental results are displayed in Table~\ref{table:Comparative Results for Different GCN Loss Functions}. 

\begin{table}[!ht]
\begin{center}
\caption{Comparative Results for Different GCN Loss Functions}
\label{table:Comparative Results for Different GCN Loss Functions}
\setlength{\tabcolsep}{1mm}{ 
\begin{tabular}{@{}c|cccccc@{}}
\toprule
Distance Function & \multicolumn{3}{c}{First Moment}                              & \multicolumn{3}{c}{Second Moment} \\ \cmidrule(r){2-4} \cmidrule(r){5-7}
Margin            & Hits@1          & Hits@5          & Hits@10 & Hits@1  & Hits@5  & Hits@10          \\ \midrule
$\gamma=3$        & 59.36          & \textbf{84.46} & 88.25  & 59.04  & 83.92  & 88.30           \\
$\gamma=5$        & \textbf{59.58} & 84.01          & 87.63  & 59.45  & 84.41  & \textbf{88.34} \\ \bottomrule
\end{tabular}}
\end{center}
\end{table}
 From Table~\ref{table:Comparative Results for Different GCN Loss Functions}, we can observe that the best Hits@1 score is obtained when $\gamma=5$ and the first moment is used. Meanwhile, Hits@5 is better when $\gamma=3$ while Hits@10 is better when the second moment is used. Overall, margin $\gamma$ and distance function affect the score in a small scale, and GCN model shows robustness in these hyper-parameters. As we pay more attention to the Hits@1, we choose $\gamma=5$ and the first moment in GCN loss function for the best Hits@1 score.

\section{Conclusion and Future Work}
\label{Sec:Conclusion and Future Work}
In this paper, we propose a novel terminology enriching method which aligns a set of terminologies to a terminology KB based on semantic embedding learned by BERT and structure embedding learned by GCN. These two embeddings are integrated to measure the relevancy of the terminology and the entity. The optimal alignment is acquired by ranking the relevancy between the terminology and all the entities in the KB. Our approach is the first one to make use of pre-trained language model to hot-start KB embedding and to enhance semantic information, and adapt GCN to our task. We evaluate our method on clinical indicator terminology KB, collected from 38 top-class hospitals of Shanghai Hospital Development Center, and experimental results show the advantages of our proposed model over the compared baselines and the ability to enrich the KB.

For future work, we plan to explore more advanced pre-trained models and GCN models to improve our proposed method. Also, we will apply our approach to other types of terminology enriching, such as disease, operation and symptom.

\section*{Acknowledgment}

We want to thank Jun Wang (Shanghai SimMed Technology Limited Company) who has helped us a lot in the construction of clinical indicator terminology KB. This work is supported by the National Key R\&D Program of China for ``Precision Medical Research" (No.~2018YFC0910500).

\bibliographystyle{IEEEtran}
\bibliography{IEEEexample}

\end{CJK*}

\end{document}